\documentclass[dvipsnames,format=sigconf]{acmart}
\usepackage[utf8]{inputenc}
\usepackage{hyphenat}
\usepackage[ruled,linesnumbered,noend,noline]{algorithm2e}
\usepackage{subcaption}
\usepackage[belowskip=-8pt,skip=5pt]{caption}

\DeclareMathOperator{\overlapping}{OverlappingEmbeddings}

\title[Exploring Layerwise Adversarial Robustness Through the Lens of t-SNE]{Exploring Layerwise Adversarial Robustness\\Through the Lens of t-SNE}

\author{Inês Valentim}
\orcid{0000-0001-6018-1788}
\affiliation{%
\institution{University of Coimbra, CISUC/LASI, DEI}
\city{Coimbra}
\country{Portugal}
}
\email{valentim@dei.uc.pt}
\author{Nuno Antunes}
\orcid{0000-0002-6044-4012}
\affiliation{%
\institution{University of Coimbra, CISUC/LASI, DEI}
\city{Coimbra}
\country{Portugal}
}
\email{nmsa@dei.uc.pt}
\author{Nuno Lourenço}
\orcid{0000-0002-2154-0642}
\affiliation{%
\institution{University of Coimbra, CISUC/LASI, DEI}
\city{Coimbra}
\country{Portugal}
}
\email{naml@dei.uc.pt}

\copyrightyear{2024}
\acmYear{2024}
\setcopyright{rightsretained}
\acmConference[GECCO '24 Companion]{Genetic and Evolutionary Computation Conference}{July 14--18, 2024}{Melbourne, VIC, Australia}
\acmBooktitle{Genetic and Evolutionary Computation Conference (GECCO '24 Companion), July 14--18, 2024, Melbourne, VIC, Australia}
\acmDOI{10.1145/3638530.3654258}
\acmISBN{979-8-4007-0495-6/24/07}


\begin{document}
\begin{abstract}
    Adversarial examples, designed to trick Artificial Neural Networks (ANNs) into producing wrong outputs, highlight vulnerabilities in these models. Exploring these weaknesses is crucial for developing defenses, and so, we propose a method to assess the adversarial robustness of image-classifying ANNs.
    The t-distributed Stochastic Neighbor Embedding (t-SNE) technique is used for visual inspection, and a metric, which compares the clean and perturbed embeddings, helps pinpoint weak spots in the layers.
    Analyzing two ANNs on CIFAR-10, one designed by humans and another via NeuroEvolution, we found that differences between clean and perturbed representations emerge early on, in the feature extraction layers, affecting subsequent classification. The findings with our metric are supported by the visual analysis of the t-SNE maps.
\end{abstract}


\maketitle

\section{Introduction}
\label{sec:introduction}
Adversarial examples~\cite{DBLP:journals/corr/GoodfellowSS14,DBLP:journals/corr/SzegedyZSBEGF13} are a threat to the robustness of Artificial Neural Networks (ANNs). They are carefully crafted to fool these models by adding perturbations, often small and imperceptible, to benign data samples~\cite{DBLP:conf/sp/Carlini017}. There is a vast literature showing that manually-designed ANNs~\cite{madry2018towards,NEURIPS2022_e1fa017a,arch-ingredients-NEURIPS2021_2bd7f907}, as well as ANNs designed in an automated way~\cite{valentim-9870202,Devaguptapu_2021_ICCV}, suffer from this vulnerability.

The robustness of a model against adversarial examples can be estimated by performing attacks and computing their success rate~\cite{croce2020robustbench}. This gives us a general idea of how the model would perform in similar conditions, but does not provide any insight into its inner workings. To tackle this, we propose a method to visualize and examine changes in the representation of the input data as it goes through the different layers of an ANN. 

Our proposal is based on the t-distributed Stochastic Neighbor Embedding (t-SNE)~\cite{JMLR:v9:vandermaaten08a} technique.
Relying on a visual analysis to quantify differences between original and altered data in many layers is a daunting task. Thus, we suggest a metric for measuring clean-perturbed data overlap in the t-SNE space.

We focused on Convolutional Neural Networks (CNNs) designed to solve image classification tasks. We inspected pre-trained models for the CIFAR-10 dataset~\cite{krizhevsky2009learning}, namely a manually-designed Wide Residual Network (WRN)~\cite{Zagoruyko2016WRN} and a CNN designed by NeuroEvolution (NE)~\cite{assuncao2018denser}, trained without any defense against adversarial perturbations. Following recent works~\cite{croce2020robustbench}, we considered $L_2$ and $L_\infty$-robustness, using three variants of the Auto-PGD (APGD) method~\cite{pmlr-v119-croce20b} as attacks.

The metric shows that network deterioration begins in the feature extraction layers, affecting how CNNs distinguish between clean and perturbed images. This is also visible in their separation on the t-SNE maps.



The paper is organized as follows.
Section~\ref{sec:background-related-work} provides some background.
The proposed approach and metric are described in Section~\ref{sec:methodology}. Section~\ref{sec:experimental-setup} details the general setup of our experiments and Section~\ref{sec:model-inspection} presents the main findings.
Section~\ref{sec:conclusion} concludes the paper and points toward future directions.

\section{Background and Related Work}
\label{sec:background-related-work}

An adversarial example~\cite{DBLP:journals/corr/GoodfellowSS14,DBLP:journals/corr/SzegedyZSBEGF13} is an input similar to a valid data point to which a model gives a highly different prediction~\cite{DBLP:journals/corr/SzegedyZSBEGF13}.
In the image domain, it is common to add small $L_{p}$-norm perturbations, bounded by a budget $\epsilon$, to the benign sample~\cite{carlini2019evaluating}. 
An attack can cause a misclassification of a sample as a specific class (targeted) or as any class as long as it is not the right one (untargeted)~\cite{DBLP:conf/sp/Carlini017}.

AutoAttack~\cite{pmlr-v119-croce20b} is an ensemble of white-box and black-box attacks that can be used as a heuristic evaluation method of the adversarial robustness of a model. 
It is adopted by the RobustBench~\cite{croce2020robustbench} benchmark, which uses standardized evaluation methodologies to keep track of the progress made in adversarial robustness.
The APGD method~\cite{pmlr-v119-croce20b}, a variation of the Projected Gradient Descent (PGD) method~\cite{madry2018towards} used by AutoAttack, progressively reduces the step size in an automated way, based on how the optimization is proceeding. 
\citeauthor{pmlr-v119-croce20b}~\cite{pmlr-v119-croce20b} also propose the Difference of Logits Ratio (DLR) loss, an alternative to the cross-entropy (CE) loss that is both invariant to shifts of the logits and to rescaling.




\citeauthor{NEURIPS2022_e1fa017a}~\cite{NEURIPS2022_e1fa017a} inspect the layerwise representations of CNNs using representation similarity metrics.
The authors investigate the similarity between the representations of clean and adversarially perturbed images, which closely relates to our work. Their findings suggest that this similarity score is typically high in earlier layers of the networks and, for undefended models, gets close to zero once the final layer is reached. This work does not include any method to visualize the representations themselves.

\section{Methodology}
\label{sec:methodology}

\begin{figure}[t]
    \captionsetup{belowskip=-10pt}
    \centering
    \includegraphics[width=\columnwidth]{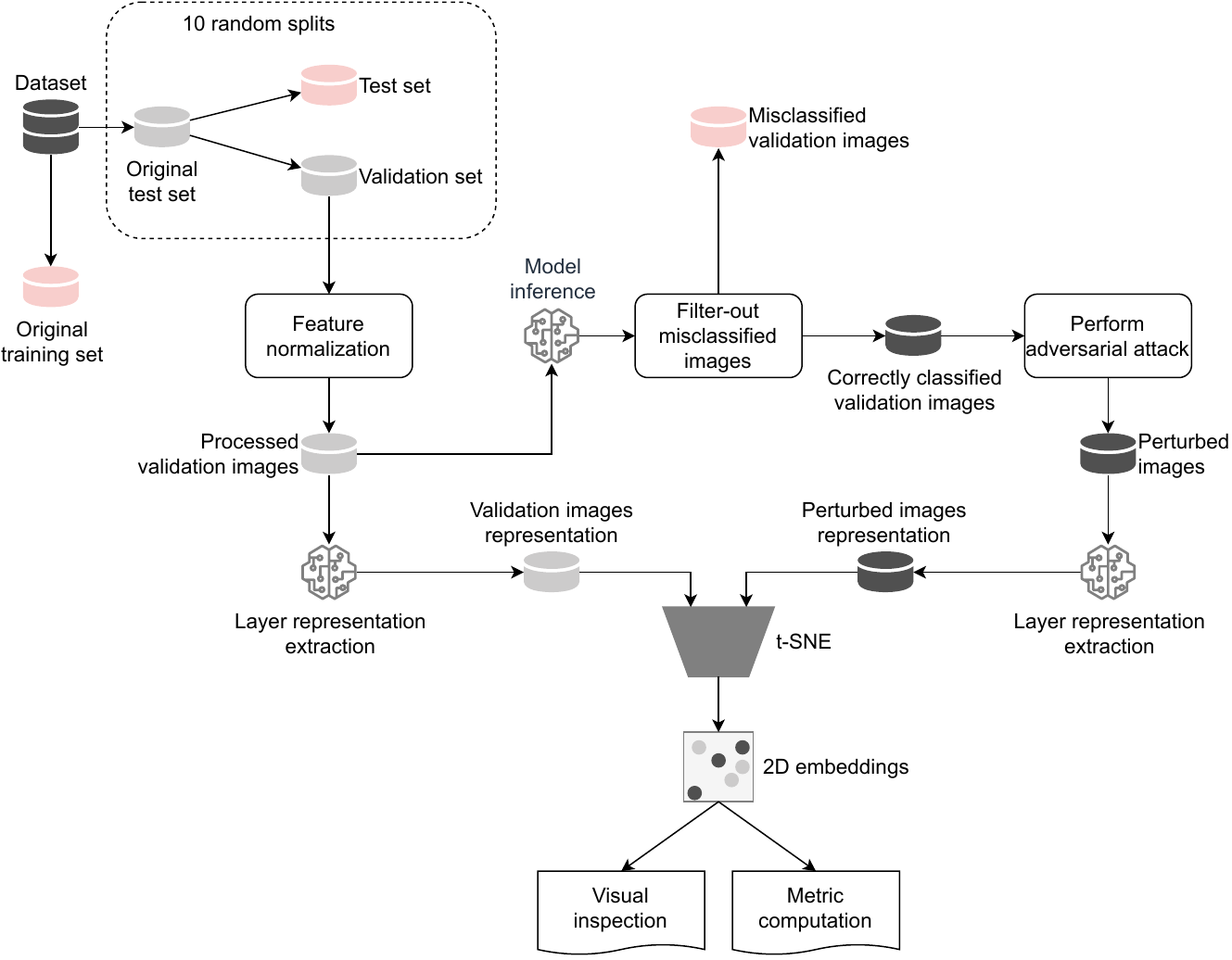}
    \caption{Proposed methodology to analyze the adversarial robustness of different layers of a CNN.}
    \label{fig:methodology}
\end{figure}

The proposed methodology to analyze the different layers of a CNN from an adversarial robustness perspective is presented in Figure~\ref{fig:methodology}.
We do not use the training set of the dataset while analyzing a model. Moreover, we create several random splits from the test set so as to be able to have access to validation data (used to analyze the model), while also putting aside test data.

The first step is to select and perform an adversarial attack, only considering correctly classified images. 
Pixel values are normalized beforehand and any pre-processing specific to a model is included in its definition.
Once the attack is applied, both the perturbed images and the clean ones are passed through the ANN up until the desired target layer to extract the hidden representations.

Due to being high-dimensional, it is not possible to visualize this latent data. Thus, for each layer that we want to inspect, we apply the t-SNE method to the extracted representations to get a two-dimensional map. We also apply the t-SNE method to the clean images incorrectly classified by the model (not used as inputs to the adversarial attacks).

Generating adversarial examples occurs once per validation set, but extracting latent layer representations and computing the t-SNE map must be repeated for each inspected layer. For each validation set and target layer, we examine the 2D maps and compute a metric that measures layer robustness by comparing clean and perturbed image embeddings.

\subsection{Robustness Metric}

To summarize the layer's outputs, we propose a robustness metric based on the differences between the clean and the corresponding adversarially perturbed representations on the t-SNE map. 
The rationale behind the proposed metric is based on the notion that, for a representation learned by a layer to be robust, the clean image and the adversarially perturbed one should be mapped to the same point in the t-SNE space.

We restrict the computation of the metric to pairs of clean-perturbed images, even though the t-SNE technique is also applied to clean images for which adversarial attacks are not generated. 
For each analyzed layer, we calculate the t-SNE for both clean and attacked images. We find the shortest Euclidean distance between each clean image and a clean instance from a different class. If the distance between a clean image and its perturbed counterpart is less than this minimum, we consider that they overlap.

The final metric value corresponds to the ratio of clean-perturbed pairs that overlap according to that heuristic:
$$\text{robustness metric} = \frac{\sum_{i=1}^{n} \overlapping\left(i, C, A, Y\right)}{n}$$
%
where $\overlapping$ is the algorithm described to detect overlaps, $C$ and $A$ are the t-SNE representations of the clean and adversarially perturbed images, respectively, $Y$ is the set of true labels of the clean images, and $n$ is the number of perturbed images that were generated (and, thus, the number of clean-perturbed pairs). 
Metric values range from 0 to 1, with higher values indicating that more clean and perturbed images overlap on the t-SNE map, which suggests layerwise robustness. This metric requires attacking instances from multiple classes, as detecting overlaps depends on the minimum distance to an instance of a different class.

\section{Experimental Setup}
\label{sec:experimental-setup}
All experiments were run in Python 3.8, using Tensorflow 2.5.0 and PyTorch 1.10.1.

\subsection{Dataset and Models}
Using 10 random seeds and maintaining data balance through stratification, the test set of the CIFAR-10 dataset~\cite{krizhevsky2009learning} was splitted into a validation set and a final test set, each with 5000 images. Adversarial examples were generated for validation images.


Given the remarkable results achieved by some NeuroEvolutionary approaches~\cite{10.1145/3321707.3321729,assuncao2018denser}, we included a CNN designed by NE~\cite{assuncao2018denser} in our experiments. Additionally, we used a handcrafted architecture~\cite{Zagoruyko2016WRN} as a baseline, specifically the WRN-28-10 model\footnote{https://github.com/RobustBench/robustbench/tree/master/robustbench/model\_zoo} trained by the RobustBench team. Regarding DENSER~\cite{assuncao2018denser}, we chose the best performing architecture\footnote{https://github.com/fillassuncao/denser-models/tree/master/CIFAR-10/net\_1} over the original evolutionary runs. To avoid introducing bias from our end, we used the pre-trained models directly, without any form of re-training.

\subsection{Threat Models, Attacks, and t-SNE}
We performed white-box attacks, since an attacker can easily have full access to the models.
Furthermore, we considered $L_{\infty}$ perturbations with $\epsilon = 8 / 255$, as well as $L_{2}$ perturbations with $\epsilon = 0.5$~\cite{croce2020robustbench}. 

Running the complete AutoAttack ensemble~\cite{pmlr-v119-croce20b} on undefended models would be unnecessarily expensive.
Thus, we performed some of the attacks from the ensemble in isolation: an untargeted APGD on the CE loss (APGD-CE), an untargeted APGD on the DLR loss (APGD-DLR), and an APGD on the targeted DLR loss with 9 target classes (denoted by APGD-T). The number of iterations for all the attacks is 100 and neither performs random restarts.

These attacks operate over the logits.
As such, a slight modification had to be introduced in the definition of the DENSER model, whose original architecture has the softmax activation directly incorporated in the last fully-connected (FC) layer. We used the original Python implementations\footnote{https://github.com/fra31/auto-attack} of the attacks.

For visualization, we used a t-SNE implementation which relies on Barnes-Hut approximations of the gradient~\cite{JMLR:v15:vandermaaten14a}. We considered the default value of $0.5$ for the parameter that controls the trade-off between speed and accuracy.
Instead of randomly initializing the solution (i.e., the two\hyp{}dimensional embedding), the PCA method~\cite{Hotelling1933} is applied to the input data to get the initial low-dimensional representation. We considered 1500 iterations and a perplexity of 50~\cite{JMLR:v9:vandermaaten08a}. 




\section{Inspecting the Networks}
\label{sec:model-inspection}


We attack only correctly classified images, leading to varying attack counts per model. The reported post-attack accuracy includes all validation images, covering both generalization errors and adversarial robustness. This promotes a fair comparison between models.



The WRN-28-10 model has a mean accuracy of 94.84\% $\pm$ 0.25\%, which is slightly higher than that of DENSER (mean accuracy of 93.68\% $\pm$ 0.31\%). This refers to the accuracy of the pre-trained models on the validation sets of clean images from CIFAR-10.

For both models and robustness scenarios, the accuracy drops to near-zero values after performing either of the three attacks (APGD-CE, APGD-DLR, or APGD-T).
Considering $L_2$ perturbations, not all validation images can be successfully perturbed with APGD-CE and APGD-DLR, but the accuracy is always below 0.45\%, showing that neither model is robust against any of these attacks.

\subsection{Robustness Metric across Layers}

\begin{figure}[th]
    \captionsetup{belowskip=0pt}
    \centering
    \begin{subfigure}{0.70\columnwidth}
        \centering
        \includegraphics[width=\columnwidth]{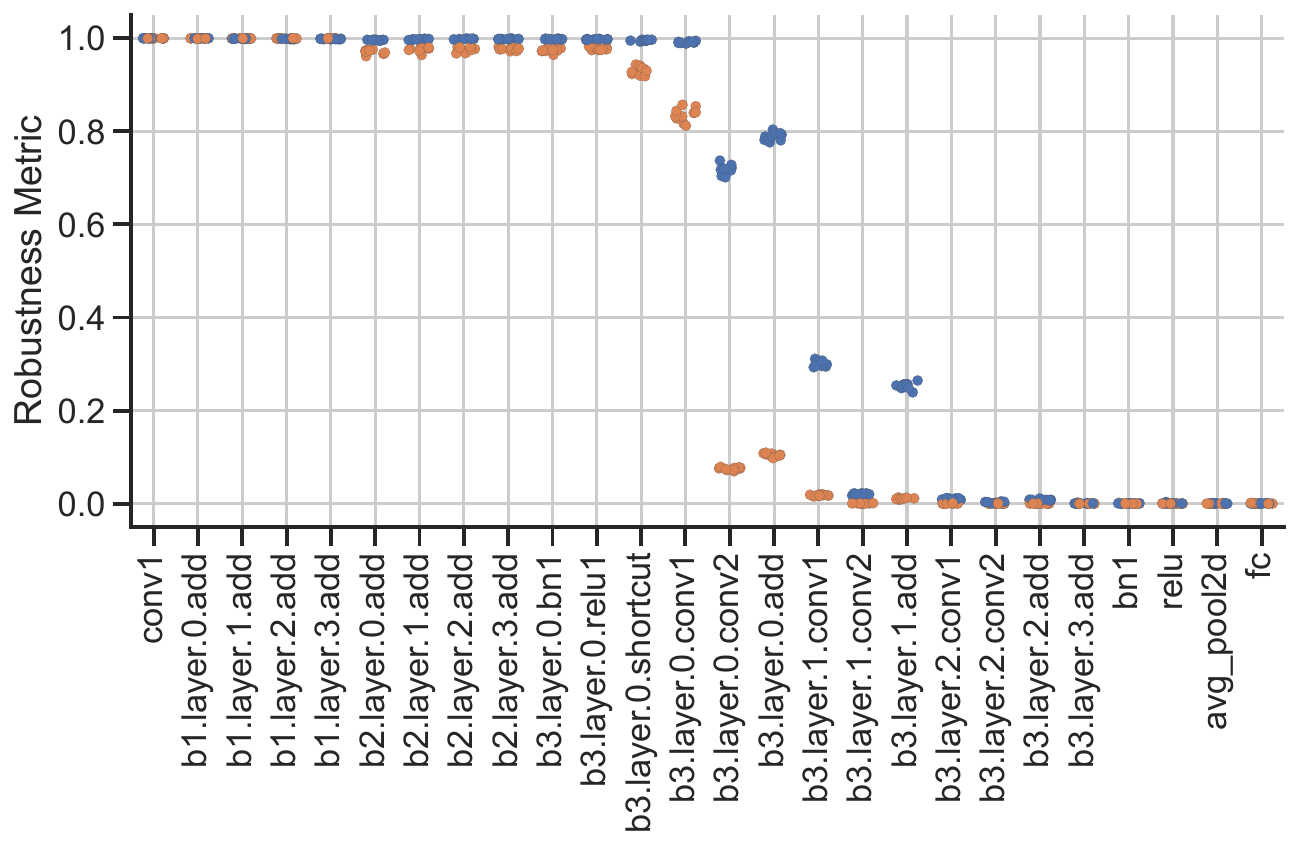}
        \subcaption{WRN-28-10.}
        \label{fig:metric-apgdce-wrn}
    \end{subfigure}
    \begin{subfigure}{0.70\columnwidth}
        \centering
        \includegraphics[width=\columnwidth]{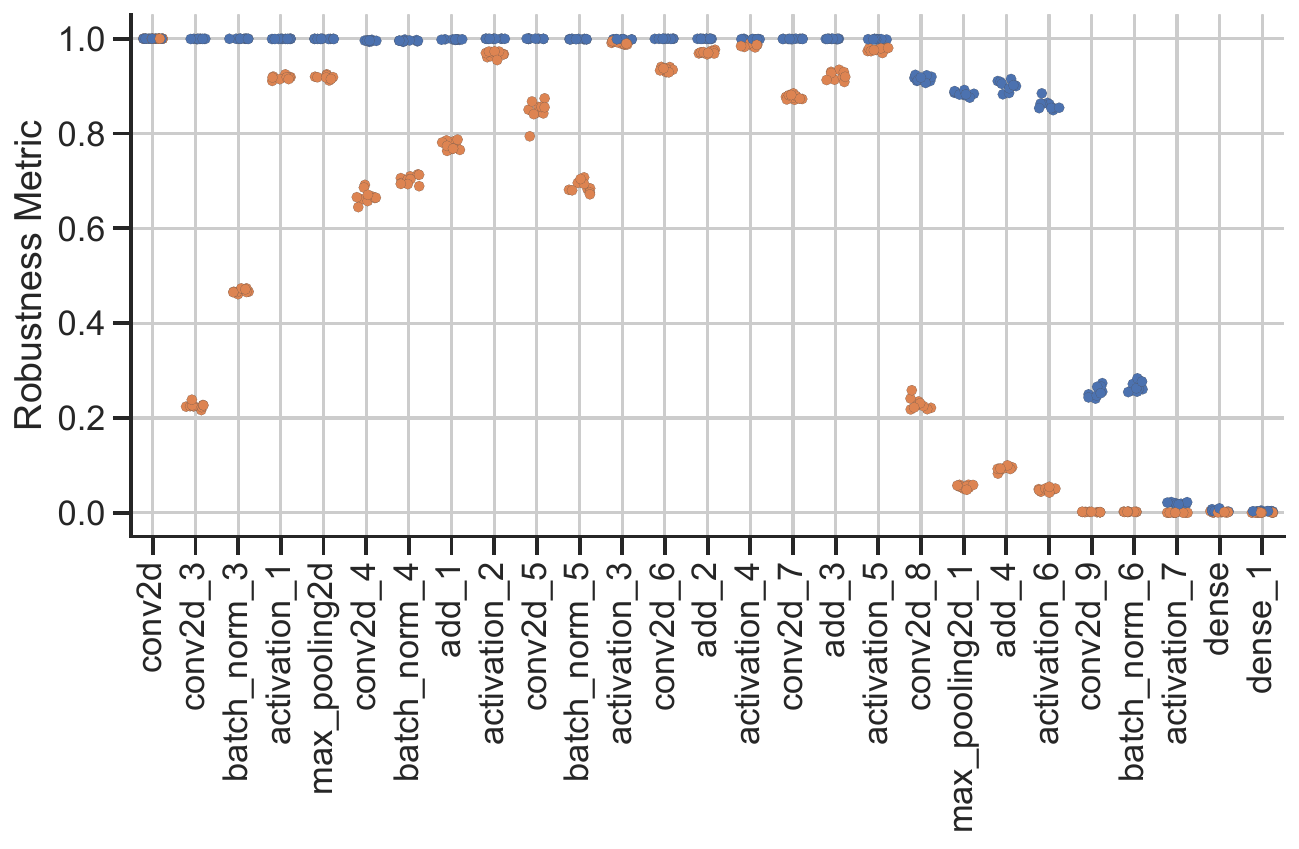}
        \subcaption{DENSER.}
        \label{fig:metric-apgdce-denser}
    \end{subfigure}
    \captionsetup{belowskip=-8pt}
    \caption{Layerwise robustness of the models under APGD-CE with $L_2$ (blue) and $L_\infty$ (orange) perturbations.}
    \label{fig:metric-apgdce}
\end{figure}

Next, we computed the robustness metric for some of the layers of the models. Figure~\ref{fig:metric-apgdce} shows the results for APGD-CE, with each point representing a run with one of the 10 validation sets. The WRN-28-10 model comprises three groups (denoted by \texttt{b1}, \texttt{b2}, and \texttt{b3}) of four residual blocks (from \texttt{layer.0} to \texttt{layer.3}). As such, \texttt{b3.layer.1.add} corresponds to the output of the second residual block in the last group. Figure~\ref{fig:metric-apgdce-wrn} shows the results for that model, while Figure~\ref{fig:metric-apgdce-denser} presents the results for DENSER.

The first layers of both models seem to keep the representations of clean and perturbed samples close to one another, as shown by metric values close to 1. Moreover, the robustness degrades earlier on in the network with $L_\infty$ perturbations than with $L_2$.

For WRN-28-10, the most drastic drops in the robustness of the hidden representations seem to occur in the last group of residual blocks. In the $L_2$ scenario, the metric drops to almost zero after the second residual block in that group, while in the $L_\infty$ case, it is almost zero for all layers following the first residual block. Metric values close to zero mean that there is practically no overlap between clean and perturbed images on the low-dimensional t-SNE map.

Regarding DENSER, the layers between \texttt{conv2d} and \texttt{conv2d\_3} are omitted since the metric does not drop from 1. In the $L_2$ scenario, the layerwise robustness starts decreasing after \texttt{activation\_5}, but the biggest drop occurs after \texttt{activation\_6}. It only gets close to zero with the last ReLU activation (i.e., \texttt{activation\_7}).
For $L_\infty$ perturbations, the results deviate from what we have observed so far. There is a significant drop in \texttt{conv2d\_3}, but the model seems to recover, with the values for \texttt{activation\_5} being close to 1. From this point onward, the metric drops significantly again, reaching zero with \texttt{conv2d\_9} and all the following layers.

The DENSER model has considerably less convolutional layers than WRN-28-10 (10 vs. 28), but more of those layers from the latter model seem to learn representations that are less robust. Focusing on $L_2$-robustness, the metric at the 23\textsuperscript{rd} convolution of WRN-28-10 (\texttt{b3.layer.1.conv1}) drops below 0.4 and does not increase in any layers that follow, while that only happens at the last convolution of DENSER. That represents more than 20\% of the convolutional layers of WRN-28-10, but only 10\% of DENSER.

Due to space restrictions, we do not show the obtained results with the remaining two attacks, but similar trends can be observed. 
We just note that, for DENSER, the metric values remain higher until later in the model than with APGD-CE.

\subsection{Visual Inspection}

\begin{figure}[ht]
    \captionsetup{belowskip=0pt}
    \centering
    \begin{subfigure}{0.45\columnwidth}
        \centering
        \includegraphics[height=70pt]{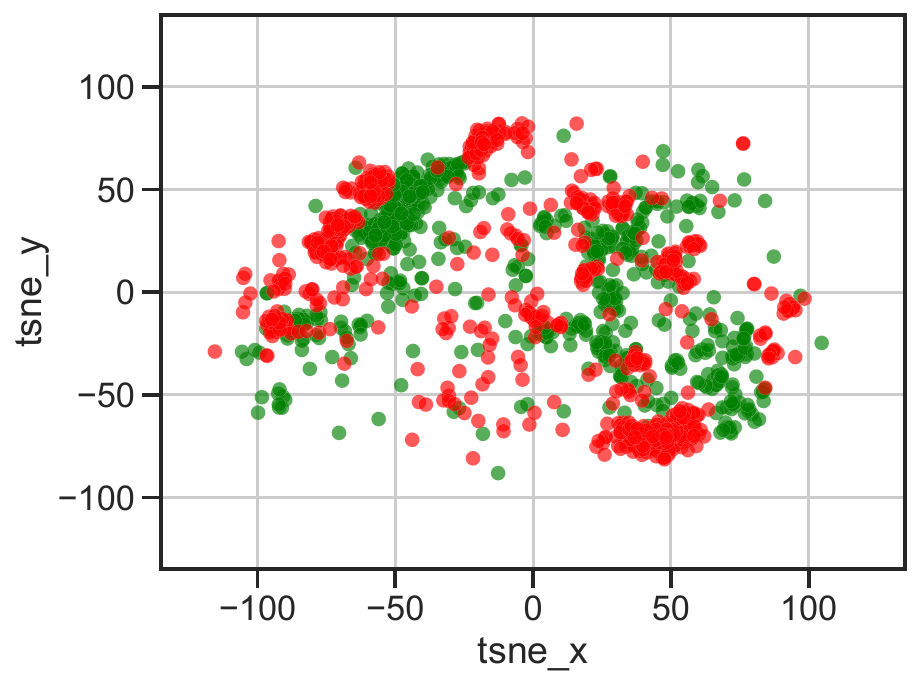}
        \subcaption{\texttt{activation\_6}.}
        \label{fig:tsne-denser-act6-l2-metric}
    \end{subfigure}\hfill
    \begin{subfigure}{0.45\columnwidth}
        \centering
        \includegraphics[height=70pt]{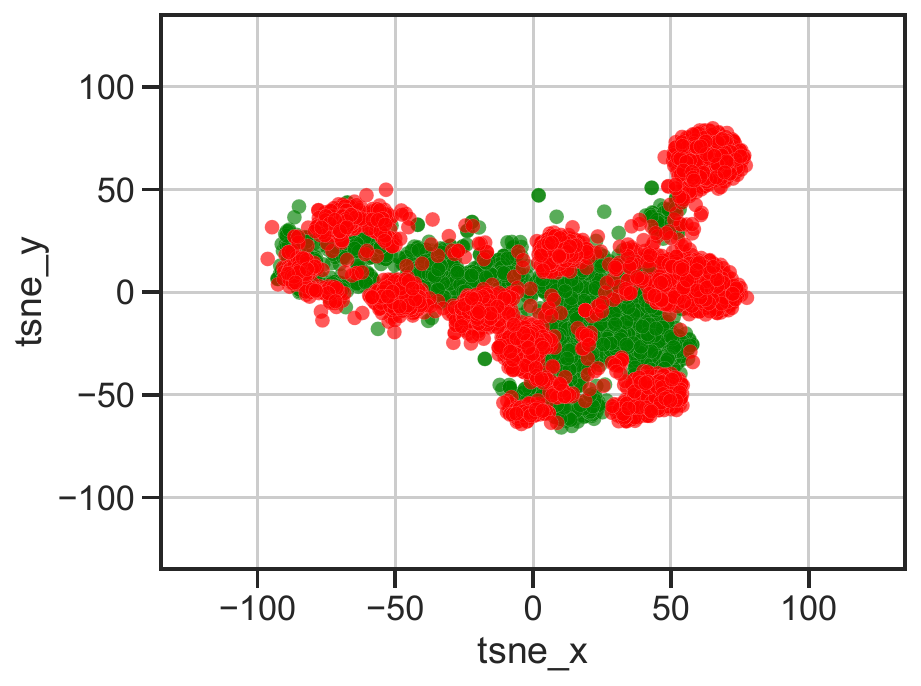}
        \subcaption{\texttt{conv2d\_9}.}
        \label{fig:tsne-denser-conv2d9-l2-metric}
    \end{subfigure}\hfill
    \captionsetup{belowskip=-8pt}
    \caption{Non-overlapping clean (green) and perturbed (red) points on the t-SNE map of different intermediate layers of the DENSER model, considering an APGD-CE attack in $L_2$.}
\end{figure}

Figure~\ref{fig:tsne-denser-act6-l2-metric} shows the 2D map for the \texttt{activation\_6} layer of DENSER, only considering the points (both clean and perturbed) that, according to the metric, do not overlap. Differences between clean and perturbed representations are relatively scarce.
These representations start to diverge in the layer that immediately follows \texttt{activation\_6}, i.e., \texttt{conv2d\_9}. Figure~\ref{fig:tsne-denser-conv2d9-l2-metric} shows the non-overlapping points for that layer, which are noticeably more than on Figure~\ref{fig:tsne-denser-act6-l2-metric}. Additionally, the perturbed points seem to encircle the clean ones.


\begin{figure}[ht]
    \captionsetup{belowskip=0pt}
    \centering
    \begin{subfigure}{0.45\columnwidth}
        \centering
        \includegraphics[height=70pt]{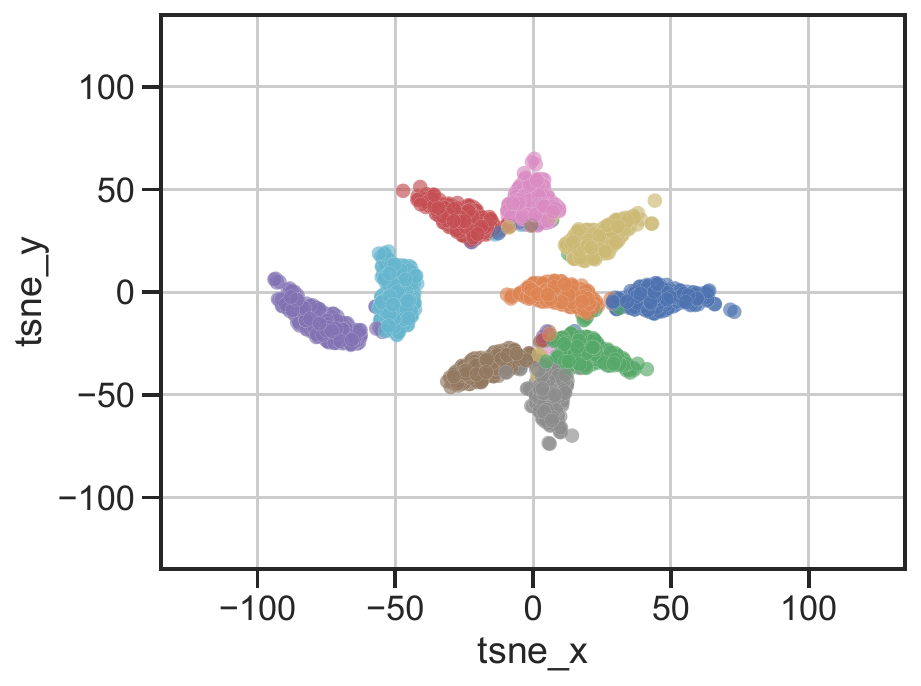}
        \subcaption{Clean.}
        \label{fig:tsne-denser-dense1-l2-clean}
    \end{subfigure}\hfill%
    \begin{subfigure}{0.45\columnwidth}
        \centering
        \includegraphics[height=70pt]{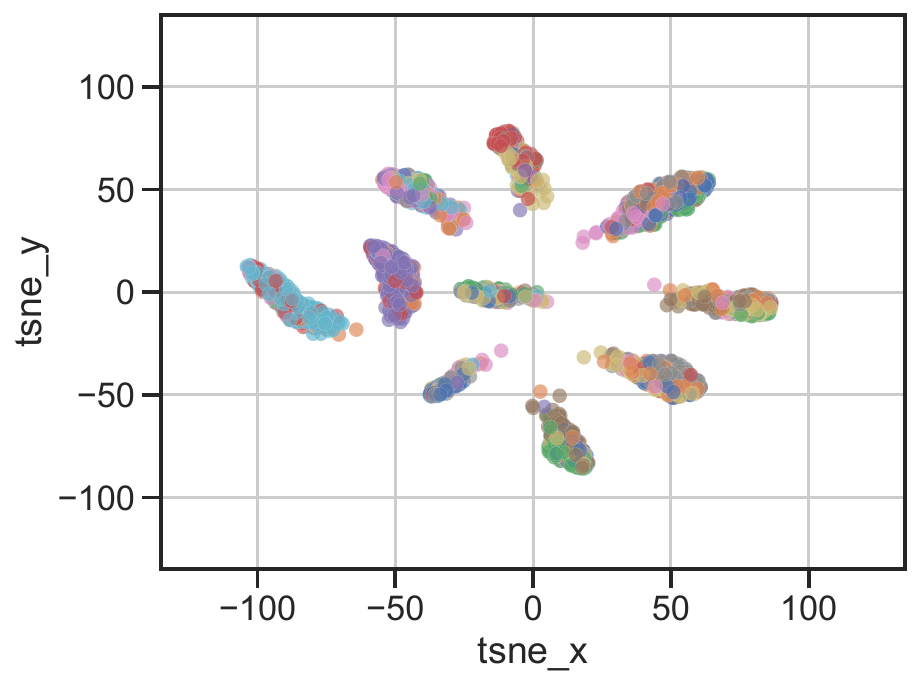}
        \subcaption{Adversarially perturbed.}
        \label{fig:tsne-denser-dense1-l2-adver}
    \end{subfigure}\hfill%
    \captionsetup{belowskip=-8pt}
    \caption{t-SNE map for the last layer of DENSER and an APGD-CE attack in $L_2$. Colors represent true labels.}
    \label{fig:tsne-denser-dense1-l2}
\end{figure}

Lastly, Figures~\ref{fig:tsne-denser-dense1-l2-clean} and~\ref{fig:tsne-denser-dense1-l2-adver} show the representation of clean and perturbed images at the final FC layer of DENSER, before softmax. Clean images cluster by true labels, while perturbed images cluster with instances from different classes. Almost no clean-perturbed points overlap once they reach this layer.


The t-SNE maps of WRN-28-10 layers further validate our metric. For instance, the map of the \texttt{b3.layer.0.add} layer resembles the one obtained for the \texttt{activation\_6} layer of DENSER. In the last layer, clean points once again cluster by class and do not overlap with perturbed ones, which are more scattered in the space but still form clusters of mixed labels.

\section{Conclusion and Future Work}
\label{sec:conclusion}

Adversarial examples compromise ANN robustness. Since traditional evaluations fall short in multi-layer analysis, we propose a method that quantifies and visually examines the discrepancies between the latent representations of clean and adversarial samples.


Our results show that discrepancies between clean and perturbed data appear still during feature extraction, even before the final convolutional layer. Our layerwise robustness metric aids defense development, with potential uses in improving NE fitness functions or selecting layers for detection-based defenses. 



For each architecture, we used a single pre-trained model, which may raise questions on generalizability. Attempts to retrain the models and reproduce the results with the pre-trained WRN were unsuccessful, highlighting the potential influence of the learning strategy on the adversarial robustness of a model.


In the future, the proposed approach needs to be evaluated on more datasets, and with models that have been explicitly designed to be adversarially robust.


\begin{acks}
This work is funded by the \grantsponsor{FCT}{FCT - Foundation for Science and Technology, I.P./MCTES}{} through national funds (PIDDAC), within the scope of CISUC R\&D Unit - \grantnum{FCT}{UIDB/00326/2020} or project code UIDP/00326/2020. It is also supported by the \grantsponsor{PRR}{Portuguese Recovery and Resilience Plan (PRR)}{} through project \grantnum{PRR}{C645008882-00000055}, Center for Responsible AI. The first author is also funded by the FCT under the individual grant \grantnum[https://doi.org/10.54499/UI/BD/151047/2021]{FCT}{UI/BD/151047/2021}.
\end{acks}

\bibliographystyle{ACM-Reference-Format}
\bibliography{references}

\end{document}